\documentclass[sigconf]{acmart}

\usepackage{caption}
\usepackage{subcaption}
\usepackage{tabularx,booktabs,multirow}
\usepackage{color}
\usepackage{balance}

\copyrightyear{2020}
\acmYear{2020}
\setcopyright{iw3c2w3}
\acmConference[WWW '20]{Proceedings of The Web Conference 2020}{April 20--24, 2020}{Taipei, Taiwan}
\acmBooktitle{Proceedings of The Web Conference 2020 (WWW '20), April 20--24, 2020, Taipei, Taiwan}
\acmPrice{}
\acmDOI{10.1145/3366423.3380015}
\acmISBN{978-1-4503-7023-3/20/04}

\begin{document}

\title{Review-guided Helpful Answer Identification in E-commerce}\thanks{The work described in this paper is substantially supported by grants
from the Research Grant Council of the Hong Kong Special Administrative
Region, China (Project Codes: 14204418) and the Direct Grant of the Faculty of
Engineering, CUHK (Project Code: 4055093).}

\author{Wenxuan Zhang, Wai Lam, Yang Deng, Jing Ma}
\affiliation{%
  \institution{The Chinese University of Hong Kong}
}
\email{{wxzhang, wlam, ydeng, majing}@se.cuhk.edu.hk}

\begin{abstract}
  Product-specific community question answering platforms can greatly help address the concerns of potential customers. However, the user-provided answers on such platforms often vary a lot in their qualities. Helpfulness votes from the community can indicate the overall quality of the answer, but they are often missing. Accurately predicting the helpfulness of an answer to a given question and thus identifying helpful answers is becoming a demanding need. Since the helpfulness of an answer depends on multiple perspectives instead of only topical relevance investigated in typical QA tasks, common answer selection algorithms are insufficient for tackling this task. In this paper, we propose the Review-guided Answer Helpfulness Prediction (RAHP) model that not only considers the interactions between QA pairs but also investigates the opinion coherence between the answer and crowds' opinions reflected in the reviews, which is another important factor to identify helpful answers. Moreover, we tackle the task of determining opinion coherence as a language inference problem and explore the utilization of pre-training strategy to transfer the textual inference knowledge obtained from a specifically designed trained network. 
  Extensive experiments conducted on real-world data across seven product categories show that our proposed model achieves superior performance on the prediction task.
\end{abstract}


\maketitle

\fancyhead{}

\section{Introduction}

\begin{figure}[h]
\setlength{\abovecaptionskip}{2pt}   
\setlength{\belowcaptionskip}{2pt}
  \centering
  \setlength{\belowcaptionskip}{-4.5mm}
  \includegraphics[width=\linewidth]{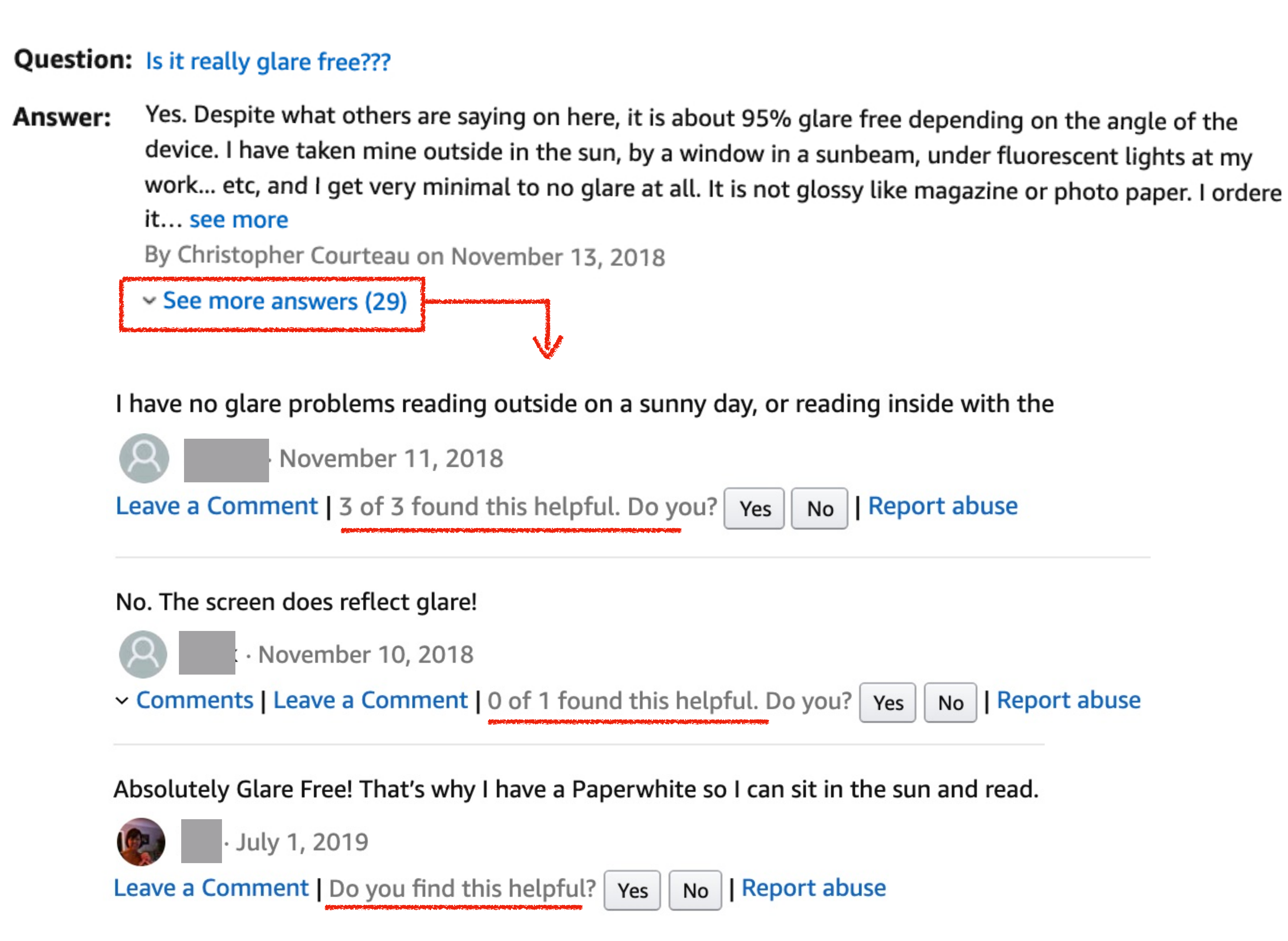}
  \caption{Example of multiple answers to a question}
  \label{multi-a}
\vspace{-0.2cm}
\end{figure}

During online shopping, a user often has some questions regarding the concerned product. To help these potential customers, product-specific community question answering (PCQA) platforms have been provided on many E-commerce sites, where users can post their questions and other users can voluntarily answer them. Unfortunately, these user-provided answers vary a lot in their qualities since they are written by ordinary users instead of professionals. Hence, they inevitably suffer from issues such as spam, redundancy, and even malicious content \cite{pqa-challenge, review-spam}.

To identify high-quality answers, many PCQA platforms allow the community to vote whether they think the answer is helpful to them or not (see Figure \ref{multi-a}, from Amazon). Such helpfulness score of each answer is often reflected in the form of $[X, Y]$, which represents that $X$ out of $Y$ users think it is helpful, where $Y$ is the total number of votes and $X$ is the number of upvotes. It serves as a vital numerical indicator for customers to get a sense of the overall quality of the answer. Such scores are also useful for E-commerce sites to recommend helpful answers to users to save their time from reading all the available ones. However, in practice, many answers do not get any vote. For instance, there are about 70\% of answers (581,931 out of 867,921) in the Electronics category without any vote at all (regardless of upvote or downvote, i.e. $X=Y=0$) in the Amazon QA dataset \cite{amazon-qa2}. This observation motivates us to investigate the task of automatic prediction of answer helpfulness in PCQA platforms, which enables the platform to automatically identify helpful answers towards the given question.

An intuitive method for such helpful answer identification task is to apply answer selection approaches as widely used for community question answering (CQA) tasks \cite{semeval-16, semeval-17, dan, DBLP:conf/coling/DengSYLDFL18}. Typically, their main goal is to determine whether a candidate answer is relevant to a question, where negative instances (i.e. irrelevant answers) are sampled from the whole answer pool \cite{wan2016deep, tay2017learning,DBLP:journals/corr/abs-1911-09801}. Since our focus is on predicting the helpfulness of the original answers written for a given question in PCQA platforms, those answers can naturally be regarded as "relevant" already. 
For example, all the three answers in Figure \ref{multi-a} are quite topically relevant to the question, but not all of them are helpful as shown in the votes they got.
Therefore, we can observe that a helpful answer is inherently relevant but not vice versa. These characteristics differentiate the helpfulness prediction task in PCQA from the CQA answer selection task by extending the quality measurement of an answer from "topically relevant to a question" to a more practically useful setting in E-commerce. 

While there are some prior works on predicting content helpfulness such as product review helpfulness \cite{review-helpfulness-survey, review-cnn} and post helpfulness \cite{post-helpful}, the answer helpfulness prediction task in E-commerce scenario has not been studied before. 
One major challenge of this task is the subjectivity of the question and answer text, indicating that even conflicting answers may exist to a specific question. 
However, we can observe that whether the opinions reflected in the answer are coherent with the common opinion regarding a specific aspect of the concerned product is an important factor for indicating the answer helpfulness. For instance, let us consider \textit{"Do most consumers agree that the Kindle Paperwhite is glare free?"} as in the example in Figure \ref{multi-a}.  This is practically meaningful since a user who bought the product before tends to upvote the answers sharing similar opinions with him/her. Such common opinions also reveal the authentic quality of that product, showing the value of the community feedback \cite{bian2008finding}. 
In E-commerce, product reviews can be such a valuable source reflecting the crowds' common opinions. Therefore, the opinion information contained in the relevant reviews can be utilized as an effective signal to guide the prediction. 

In this paper, we propose a Review-guided Answer Helpfulness Prediction (RAHP) model to tackle the helpful answer identification task. It not only considers the interactions between QA pairs, but also utilizes relevant reviews to model the opinion coherence between the answer and common opinions.
In specific, we first employ a dual attention mechanism to attend the important and relevant aspects in both the question and answer sentences.
Then the relevant reviews are utilized for analyzing the opinion coherence. We further observe that this component, in essence, can be modeled as a natural language inference (NLI) problem (i.e., recognizing textual entailment \cite{snli-dataset, xnli}). Specifically, the opinion coherence between the answer and the review can be viewed as whether the meaning of the answer ("hypothesis") can be inferred from the review ("premise"). To tackle the issue of lacking labeled data supporting the learning of such review-answer (RA) entailment, we explore the utilization of labeled language inference datasets via pre-training an appropriate neural network. Then the knowledge can be transferred to our target review-answer entailment analysis component.
The implementation for our model is publicly available at \url{https://github.com/isakzhang/answer-helpfulness-prediction}.

To sum up, our main contributions are as follows:
(1) We study a novel task, answer helpfulness prediction, to identify helpful answers in PCQA platforms. 
(2) We propose the RAHP model to incorporate review information with a textual entailment module to jointly model the relevance between the QA pairs and the opinion coherence between the answer and the reviews. 
(3) Experimental results show that our model achieves superior performance on real-world datasets across seven product categories. 
(4) We employ a pre-training strategy to transfer prior knowledge of recognizing textual entailment patterns, which further improves the performance.

\section{Related Work}
\noindent \textbf{Answer Selection in CQA.}
Given the popularity of online community forums, community question answering (CQA) has become an emerging research topic in recent years \cite{semeval-16, semeval-17, cqa-survey, cqa-question-retrieval}. One major research branch of CQA is the answer selection task which aims to select the most relevant answers from a large candidate answer set. Earlier works of answer selection task relied heavily on feature engineering such as utilizing tree matching approaches \cite{cui05}. Topic models were also employed to detect the similarity in the latent topic space \cite{cqa-topic, cqa-topic-2}. To avoid feature engineering, many deep learning models have been proposed for the answer selection task recently. \citet{qa-lstm} proposed a LSTM-based encoding method to encode the question and the answer sentence to make the prediction. They also explored the utilization of an attention mechanism. A two-way attention approach was introduced later for better sentence representations \cite{dos}. Some models with elaborately designed architecture were also proposed to carefully measure the relevance between the question and the answer. \citet{esim} utilized several matrix attention and inference layers to analyze the relevance information. \citet{cqa-acl18} exploited the specific structure of some CQA platforms and separately processed question subject and question body to improve the question representations. Although achieving good performance on answer selection task, these models fall short to measure answer helpfulness in E-commerce scenario, as the rich information in the review contents is underutilized.

\noindent \textbf{Content Helpfulness Prediction.}
A series of works have been conducted on measuring the helpfulness of text contents in different domains such as online education \cite{mooc} and discussion forums \cite{post-helpful}. 
Among them, helpfulness prediction of product reviews received a lot of attention. An up-to-date survey of various works can be found in a recent paper \cite{review-helpfulness-survey}. To analyze the helpfulness of a review text, many features are considered such as the length \cite{review-len}, the readability \cite{review-read} and the lexical features of the reviews \cite{review-word}. There are also some works exploring the utilization of deep learning based prediction, where domain knowledge are often incorporated to these models for improving the performance. \citet{review-cnn} proposed to utilize the character-level embedding to enrich the word representation and tackle the domain knowledge transfer. 
\citet{prh-net} proposed a neural architecture fed by both the content of a review and the title of the concerned product to conduct the product-aware prediction.

\section{Our Proposed Model}
\setlength{\abovecaptionskip}{2pt}   
\setlength{\belowcaptionskip}{2pt}
\begin{figure*}[h]
  \centering
  \includegraphics[width=160mm]{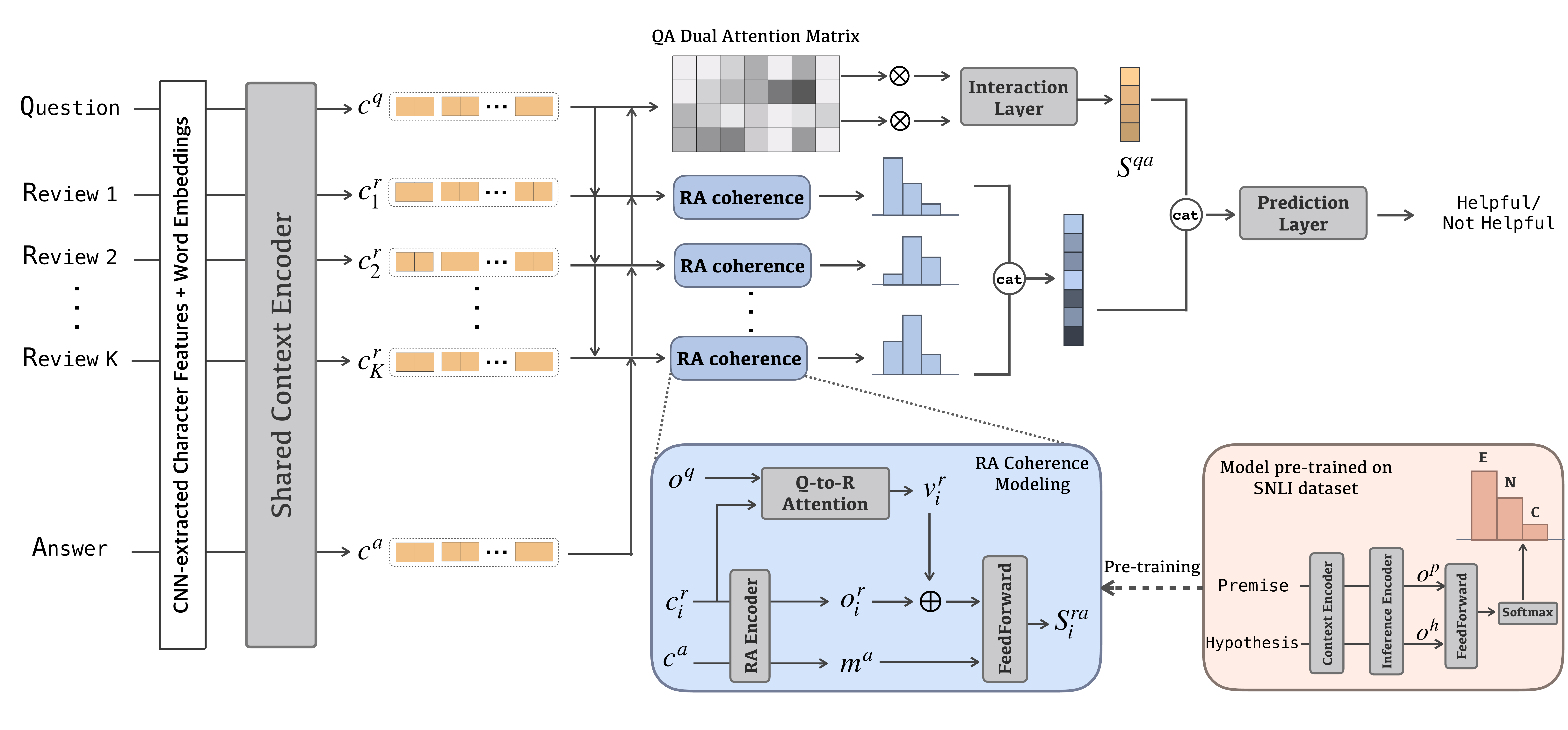}
  \caption{The architecture of our proposed RAHP model and a Siamese model for textual inference pre-training}
  \label{model}
  \vspace{-0.4cm}
\end{figure*}

In this section, we describe our proposed Review-guided Answer Helpfulness Prediction model (RAHP) for identifying helpful answers in PCQA. Given an answer $a$, the corresponding question $q$ and a set of $K$ relevant review sentences $\{r_1, r_2, ..., r_K\}$, RAHP aims at predicting whether $a$ is a helpful answer or not. As shown in Figure \ref{model}, it is mainly composed of three components: 1) QA interaction modeling, 2) review-answer (RA) coherence modeling and 3) knowledge transfer from a pre-trained textual inference network.

\subsection{Context Encoding}
We denote the sequence length of the answer and the question as $L_a$, $L_q$ respectively. For the \textit{i}-th review, its sequence length is denoted as $L_{r_i}$. We first embed each word of them into a low-dimensional dense vector $e(w)=[e_{c}(w); e_{g}(w)]$ as a concatenation of character-level embedding $e_{c}(w)$ and word embedding $e_{g}(w)$, where $e_{c}(w)$ is learned from a convolutional neural network~\cite{cnn-kim} and $e_{g}(w)$ is the pre-trained Glove word vectors \cite{glove}. We use $[ \cdot \ ; \cdot ]$ to denote the concatenation operation.
After transforming each word into a vector representation, we employ a bidirectional LSTM \cite{bilstm} to capture the local context information:
\begin{equation} \label{bilstm.c}
    h_t=\operatorname{BiLSTM}^{c}(e(w_t), h_{t-1})
\end{equation}
where $h_t$ denotes the hidden state of the BiLSTM at the \textit{t}-th time step, $w_t$ is the \textit{t}-th word in the corresponding sequence. To make the presentation more clear, we will use different superscripts to discriminate different modules. For example, $c$ here in Eq. \ref{bilstm.c} indicates that this BiLSTM is the context encoding module. We then take the output of the BiLSTM at each time step as the new representation for each word. Such transformation is conducted for all inputs and the context-aware representations for the question $q$, answer $a$ and \textit{i}-th review $r_i$ are denoted as $c^q, c^a$ and $c^r_i$ respectively.

\subsection{QA Interaction Modeling}
To analyze the fine-grained interactions between the question and the answer, we employ a dual attention mechanism inspired by \cite{decompose16} to highlight the important semantic units in both the question and answer, and avoid the distractions of unimportant information. 
Since each word in the question can be similar to several words in the answer and vice versa, we then compute the similarity between a given word in the question with every other word in the answer:
\begin{equation}
    \alpha^q_j = \operatorname{softmax} (\operatorname{sim}(c^q_j, c^a_1), ..., \operatorname{sim}(c^q_j, c^a_{L_a}))
\end{equation}
where $c^*_j$ denotes the context-aware representation of the \textit{j}-th word in the question/answer sentence, $\alpha^q_j \in \mathbb{R}^{L_a}$ is thus the alignment vector for the \textit{j}-th word of the question $q$. For the choice of the similarity function $\operatorname{sim}()$ between the \textit{j}-th word in $q$ and the \textit{k}-th word in $a$, the dot product operation is employed:
\begin{equation}
    \operatorname{sim}(c^q_j, c^a_k) = c^q_j \cdot c^a_k 
\end{equation}

We also tried other choices such as utilizing a bilinear layer to compute the similarity which leads a similar performance. Thus we choose the simplest form. Therefore, we can compute an answer-enhanced representation for each word in the question. Specifically, for the \textit{j}-th word of $q$, we have:
\begin{equation}
    n^{aq}_j = \sum\nolimits_{l=1}^{L_a} \alpha_{jl}^q \cdot c^a_l
\end{equation}
where $n^{aq}_j$ is the answer-enhanced representation for the \textit{j}-th word of $q$, given by a weighted sum of the answer representation. 

Similarly, we can also obtain a question-enhanced answer representation by computing the similarity of a word in the answer with every word in the question, which gives us a new representation $n^{qa}_j$ for the $j$-th word in the answer.
After conducting such dual attention operation, we get the soft alignments from both directions, bringing us the enhanced question and answer representations for better predictions, denoted as $n^{aq}$ and $n^{qa}$ respectively. 

We then concatenate the context-aware representation and the attention enhanced representation for both the question and answer and employ another BiLSTM layer to encode them into fixed-size vector representations respectively:
\begin{gather}
    o^q = \operatorname{BiLSTM}^{qa}([c^q; n^{aq}])_{Lq} \\
    o^{a} = \operatorname{BiLSTM}^{qa}([c^a, n^{qa}])_{La}
\end{gather}
where $o^q$ and $o^a$ are the encoded representations for the question and answer respectively, taken from the final hidden state of the BiLSTM. Finally, we concatenate these two encoded representations and feed them into a MLP layer to get a low-dimensional helpfulness prediction vector denoted as $s^{qa}$:
\begin{equation}
   s^{qa} = \operatorname{MLP}^{qa}([o^q; o^a])
\end{equation}
where $s^{qa}\in \mathbb{R}^{d_1}$, $d_1$ is the dimension of this prediction vector.

\subsection{Review-Answer (RA) Coherence Modeling}
Investigating whether the opinions in the answer are coherent with the common opinions reflected in the reviews can be another important signal for the helpfulness prediction. Thus, we first employ another BiLSTM to encode the context-aware answer and review representation $c^a$ and $c^r_i$ as follows:
\begin{equation}
    m^{a} = \operatorname{BiLSTM}^{ra}(c^a)_{L_a} \; \; \; o^r_i = \operatorname{BiLSTM}^{ra}(c^r_i)_{L_{r_i}} 
\end{equation}
Similarly, we take the final hidden state of this BiLSTM to obtain vector representations for the answer and \textit{i}-th review sentence and denote them as $m^{a}$ and $o^r_i$ respectively.

However, we can observe that even for a single review, it may discuss about multiple aspects of the concerned product. This is because reviews are not originally written as a response to a specific concern, thus containing irrelevant information. To tackle this issue, we employ an attention mechanism between the question and review ("Q-to-R attention") to capture the salient information in the review. Then for the \textit{j}-th word in the \textit{i}-th review, we have:
\begin{gather}
    u_{ij}^r = o^q \cdot c^r_{ij} \\
    \beta^r_{ij} = \operatorname{exp}(u_{ij}^r) / \sum\nolimits_{l=1}^{L_{r_i}} \operatorname{exp}(u_{il}^r)
\end{gather}
where $c^r_{ij}$ is the context representation of the \textit{j}-th word in the \textit{i}-th review $r_i$, $u_{ij}^r$ and $\beta^r_{ij}$ can be regarded as the raw and normalized association measure of the \textit{j}-th word in the review to the whole question sentence. Therefore, we can obtain a question-attended review representation $v^r_i$ as:
\begin{equation}
    v^r_i = \sum\nolimits ^{L_{r_i}}_{l=1} \beta^r_{il} \cdot c^r_{il}
\end{equation}

To combine the review representations from the RA entailment encoding module and attention operation from the question, we conduct an element-wise summation:
\begin{equation}
    m^r_i = v^r_i \oplus o^r_i
\end{equation}
where $m^r_i$ is the new composite representation for the \textit{i}-th review, $\oplus$ denotes the element-wise summation. Finally, we concatenate the review representation $m^r_i$ and answer representation $m^{a}$ and pass them into a fully-connected layer to get a low-dimensional prediction vector:
\begin{equation}
    s^{ra}_i = \operatorname{MLP}^{ra}([m^r_i; m^a]) \in \mathbb{R}^{d_2}
\end{equation}

For $K$ available relevant reviews, we conduct similar operations introduced above to get the prediction vector $s^{ra}_1,...,s^{ra}_K$ respectively. Then they are concatenated with the prediction vector $s^{qa}$ obtained from the analysis of QA interactions and fed into a final MLP classifier to predict the helpfulness of the answer:
\begin{equation}
    \hat{y} = \operatorname{MLP}^p ([s^{qa}; s^{ra}_1;...;s^{ra}_K])
\end{equation}
where $[s^{qa}; s^{ra}_1;...;s^{ra}_K] \in \mathbb{R}^{d_1+K\cdot d_2}$ and $\hat{y}$ denotes the final prediction given by the model.

\subsection{Inference Knowledge Transfer}
The review-answer (RA) entailment analysis component introduced in the last section attempts to investigate the entailment relationship between an answer and a relevant review. One issue of such component is the lack of explicit supervision signal of recognizing textual inference patterns, resulting in difficulty for the prediction. To tackle this challenge, we utilize some existing language inference datasets with explicit labels to obtain some prior knowledge. Specifically, the knowledge of recognizing entailment relations in the trained model can be transferred to our target component.

We utilize the widely-used Stanford Natural Language Inference dataset (SNLI)  \cite{snli-dataset} to pre-train the network. It has three types of labels, namely, entailment, neutral, and contradiction. As shown in the right bottom part of Figure \ref{model}, we construct a similar network architecture given two input sentences in SNLI: premise and hypothesis. First, the words are embedded into vector representations, followed by a context encoding module $\operatorname{BiLSTM}^c$ to obtain the context-aware representations for premise and hypothesis, denoted as $c^p$ and $c^h$ respectively. Next, another BiLSTM is utilized to encode the premise and hypothesis into fixed-size vector representations, denoted as $o^p$ and $o^h$. The encoded representations are then concatenated together to make the final predictions $\hat{y}_{hp}$:
\begin{gather}
        o^{p} = \operatorname{BiLSTM}^{ra}(c^p)_{L_p} \; \; \; o^h = \operatorname{BiLSTM}^{ra}(c^h)_{L_h} \\
    \hat{y}_{hp} = \operatorname{MLP}^{ra} ([o^p; o^h])
\end{gather}

After training on the SNLI dataset, the trained modules capture some knowledge of recognizing the entailment patterns. For example, the text encoding module $\operatorname{LSTM}^{ra}$ now learns to capture some major information relevant for the final prediction during the encoding phase. Thus, we utilize the learned parameters of these pre-trained modules to initialize the parameters of our RA coherence modeling component. Specifically, the parameters of the context encoding module $\operatorname{BiLSTM}^c$, the inference encoding module $\operatorname{BiLSTM}^{ra}$ and the prediction module $\operatorname{MLP}^{ra}$ are transferred to RAHP for providing the prior knowledge of recognizing inference patterns.

\begin{table}
\setlength{\abovecaptionskip}{2pt}%
 \setlength{\belowcaptionskip}{2pt}
  \small
  \caption{Overview of the datasets}
  \label{dataset}
  \begin{tabular}{cccccc}
    \toprule
    Category & \# products & \# QA  & \# reviews & $L_q$ & $L_a$ \\
    \midrule
    Electronics & 17,584 & 53,514  & 657,345 & 13.4 & 16.9 \\
    Sports & 7,609 & 23,337  & 187,996 & 12.5 & 17.3 \\
    Health & 6,197 & 22,377  & 243,782  & 12.2 & 17.2\\
    Home & 12,858 & 48,441  & 489,955 & 12.4 & 17.5 \\
    Patio Lawn & 3,864 & 11,963  & 98,583 & 13.0 & 17.6 \\
    Phones & 4,022 & 11,779  & 138,615 & 12.6 & 16.0 \\
    Toys \& Games & 3,667 & 10,516  & 73,082 & 11.5 & 16.5 \\
    \bottomrule
\end{tabular}
\vspace{-0.4cm}
\end{table}

\begin{table*}
\centering
\setlength{\abovecaptionskip}{2pt}   
\setlength{\belowcaptionskip}{2pt}
\fontsize{9}{10}\selectfont
  \caption{Comparison of the F1 and AUROC scores between RAHP-Base, RAHP-NLI and comparative models}
  \label{results}
  \begin{tabular}{lcccccccccccccc}
    \toprule
      & \multicolumn{2}{c}{Electronics} 
      & \multicolumn{2}{c}{Sports}
      & \multicolumn{2}{c}{Health} 
      & \multicolumn{2}{c}{Home} 
      & \multicolumn{2}{c}{Patio Lawn} 
      & \multicolumn{2}{c}{Phones}
      & \multicolumn{2}{c}{Toy \& Games} \\
    
    \cmidrule(lr){2-3} \cmidrule(lr){4-5} \cmidrule(lr){6-7} \cmidrule(lr){8-9} \cmidrule(lr){10-11} \cmidrule(lr){12-13} \cmidrule(lr){14-15} 
    
    & F1 & AUC & F1 & AUC & F1 & AUC & F1 & AUC & F1 & AUC & F1 & AUC & F1 & AUC \\
    \midrule
    DAN & 0.604 & 0.705 & 0.589 & 0.715 & 0.603 & 0.734 & 0.636 & 0.733 & 0.622 & 0.736 & 0.561 & 0.713 & 0.586 & 0.730 \\
    QA-LSTM  & 0.597 & 0.756 & 0.587 & 0.731 & 0.581 & 0.740 & 0.666 & 0.762 & 0.603 & 0.732 & 0.581 & 0.702  & 0.550 & 0.725\\
    Att-BiLSTM & 0.622 & 0.754 & 0.604 & 0.733 & 0.566 & 0.751 & 0.701 & 0.780 & 0.640 & 0.742 & 0.571 & 0.706  & 0.610 & 0.753\\
    ESIM  & 0.623 & 0.766 & 0.632 & 0.740 & 0.643 & 0.745 & 0.701 & 0.786 & 0.644 & 0.750 & 0.586 & 0.707 & 0.556 & 0.723\\
    \midrule
    CNNCR-R & 0.596 & 0.749 & 0.624 & 0.734 & 0.590 & 0.718 & 0.704 & 0.770 & 0.653 & 0.731 & 0.581 & 0.713  & 0.560 & 0.713\\
    PH-R  & 0.586 & 0.763 & 0.614 & 0.723 & 0.631 & 0.731 & 0.645 & 0.787 & 0.649 & 0.742 & 0.569 & 0.664 & 0.515 & 0.736\\
    PRHNet-R & 0.579 & 0.746 & 0.633 & 0.733 & 0.586 & 0.724 & 0.645 & 0.761 & 0.600 & 0.713 & 0.593 & 0.712 & 0.596 & 0.720\\
    \midrule
    \midrule
    \textbf{RAHP-Base} & \textbf{0.651} & 0.770 & 0.655 & 0.753 & 0.657 & 0.765 & 0.723 & \textbf{0.801} & \textbf{0.684} & 0.761 & 0.608 & 0.720  & 0.633 & 0.754\\
    \textbf{RAHP-NLI} & 0.647 & \textbf{0.779} & \textbf{0.669} & \textbf{0.764} & \textbf{0.667} & \textbf{0.770} & \textbf{0.725} & 0.794 & 0.671 & \textbf{0.764} & \textbf{0.629} &  \textbf{0.735} & \textbf{0.636} & \textbf{0.764}\\
    \bottomrule
  \end{tabular}
\vspace{-0.3cm}
\end{table*}

\section{Experiments}
\subsection{Experimental Setup}
\subsubsection{\textbf{Datasets and Evaluation Metrics}}
We experiment with seven datasets from different product categories to validate the model effectiveness. The statistics are shown in Table \ref{dataset}. The original question-answer pairs are from a public data collection crawled by Wan and McAuley \cite{amazon-qa2}. We also utilize the product ID in the QA dataset to align with the reviews in Amazon review dataset \cite{amazon-review-dataset} so that the corresponding reviews of each product can be obtained. 

Following previous work \cite{post-helpful, prh-net}, we treat the helpful votes given by customers as a proxy of the helpfulness of each answer and model this task as a binary classification problem. 
Since users' votes are not always reliable  \cite{suryanto2009quality} and people tend to upvote when they decide to vote an answer. We discard answers with less than two votes and treat the answer to be \textit{helpful} if it receives helpfulness score being one (i.e. X/Y=1, Y$\geq$2) to obtain a high standard notion of helpfulness and a more reliable dataset.
However, we observe that answers with only one negative vote often provide reliable examples for unhelpful answers, they are kept in the dataset. 
The number of question-answer pairs available after the filtering is shown under the column "\# QA" in Table \ref{dataset}. We split the dataset in each product category into portions of 80:10:10 for training, validation, and testing respectively.

Since the class distributions are skewed among all categories, we adopt the F1 score and the Area Under Receiver Operating Characteristic (AUROC) score as the evaluation metrics.

\subsubsection{\textbf{Comparative Models}}
To evaluate the performance of our proposed model, we compare with the following strong baselines:
(1) \textbf{DAN} \cite{dan}: It adopts a Siamese architecture which encode a sentence by taking the average of word vectors, predictions are made based on the encoded sentence representation.
(2) \textbf{QA-LSTM} \cite{qa-lstm}: It employs a Siamese LSTM network to encode the question and the answer.
(3) \textbf{Attentive-BiLSTM} \cite{qa-lstm}: It improves simple LSTM by using a Bidirectional LSTM as well as an attention mechanism.
(4) \textbf{ESIM} \cite{esim}: It is one of the state-of-the-art models for the answer selection task and similar text matching problem \cite{elmo, multinli} with a complicated encoding and attention architecture.
For the content helpfulness prediction models, we also modify them to take relevant reviews, in addition to QA pairs, as their model inputs (denoted with a suffix "-R") for a more comprehensive and fair comparison:
(5) \textbf{CNNCR-R} \cite{review-cnn}: CNN with character representation is one of the state-of-the-art models to predict review helpfulness. 
We also use the same CNN encoder to encode reviews and concatenate the encoded reviews with QA pairs together.
(6) \textbf{PH-R} \cite{post-helpful}: \textbf{P}ost \textbf{H}elpfulness prediction is the state-of-the-art model for predicting whether a target post is helpful given the original post and several past posts. We treat the question and answer as the original and target post respectively and use relevant reviews to replace the several past posts. 
(7) \textbf{PRHNet-R} \cite{prh-net}: one of the state-of-the-art models for product review helpfulness prediction. We concatenate the QA pair as a "single review" and treat relevant reviews as the corresponding product information as used in the original model.

For our proposed RAHP model, we consider following two variants: \textbf{RAHP-Base}: Our proposed model RAHP with all parameters trained from scratch. \textbf{RAHP-NLI}: RAHP with the parameters of the review-answer entailment component initialized with the pre-trained network on the SNLI dataset.

\subsubsection{\textbf{Implementation Details}}
For each product, we first retrieve its all corresponding reviews and split them at the sentence level. To obtain relevant reviews for each QA pair, we utilize a pre-trained BERT model\footnote{\url{https://github.com/google-research/bert\#pre-trained-models}} \cite{bert} as the sentence encoder to find out the most relevant reviews in terms of the dot product between two vectors.\footnote{Note that any other off-the-shelf retrieval system can also be employed here if it can return a feasible set of relevant reviews for assisting the prediction.} The number of relevant reviews $K$ used in our model is set to 5. 
The word embedding matrix is initialized with pre-trained Glove vectors \cite{glove} with the dimension being 300. For the CNN-based character embeddings, the number of filters are set to be 50 and 4 types of filter with sizes \{2, 3, 4, 5\} are used. The hidden dimension of all the LSTM cell in RAHP model is set to 128. Weights of the linear layer are initialized with Xavier uniform.

\subsection{Results and Discussions}

\subsubsection{\textbf{Answer Helpfulness Prediction}} \label{5.1}

Table \ref{results} presents the results of different models over seven product categories in terms of F1 and AUROC scores (AUC) respectively. Overall, our proposed methods substantially and consistently outperform those baseline methods in all domains.
Concretely, we can observe that answer selection models generally provide strong baselines for the concerned helpfulness prediction task. Especially models with advanced architecture (e.g. ESIM model) achieves good performance due to the fact that those complicated models explicitly consider the fine-granularity relations between QA pairs. However, RAHP-Base consistently outperforms them, which demonstrates the effectiveness and necessity of taking relevant reviews into consideration.

Furthermore, comparing the performance between answer selection models and content helpfulness prediction models, it can be observed that the latter often achieves better performance among many product categories, especially in terms of F1 scores. This again may due to the reason that we augment review information into these models to help guide the prediction, thus leading to some performance improvements. Comparing RAHP-Base with these content helpfulness prediction models, we can observe that RAHP-Base can still consistently achieve better performance. Although the comparative content helpfulness prediction models are already modified to let them consider relevant reviews, they cannot explicitly exploit the interactions between these information sources. For example, we modify the review helpfulness prediction model CNNCR \cite{review-cnn} to let it also encode reviews with the same CNN encoder, but it lacks the ability to explicitly consider the opinion coherence between the answer and those relevant reviews. This observation demonstrates that the special design for measuring the opinion coherence between the reviews and answers is beneficial. Moreover, we can see that the performance on a majority of domains can be further improved with the help of pre-training on the language inference dataset (i.e. RAHP-NLI), which shows that such pre-training approach can effectively transfer some prior textual inference knowledge to the review-answer coherence modeling component, leading to more accurate helpfulness predictions. 

\subsubsection{\textbf{Effectiveness of Pre-training}}
\begin{figure}
  \centering
  \setlength{\abovecaptionskip}{2pt}
  \setlength{\belowcaptionskip}{2pt}
  \includegraphics[width=62mm]{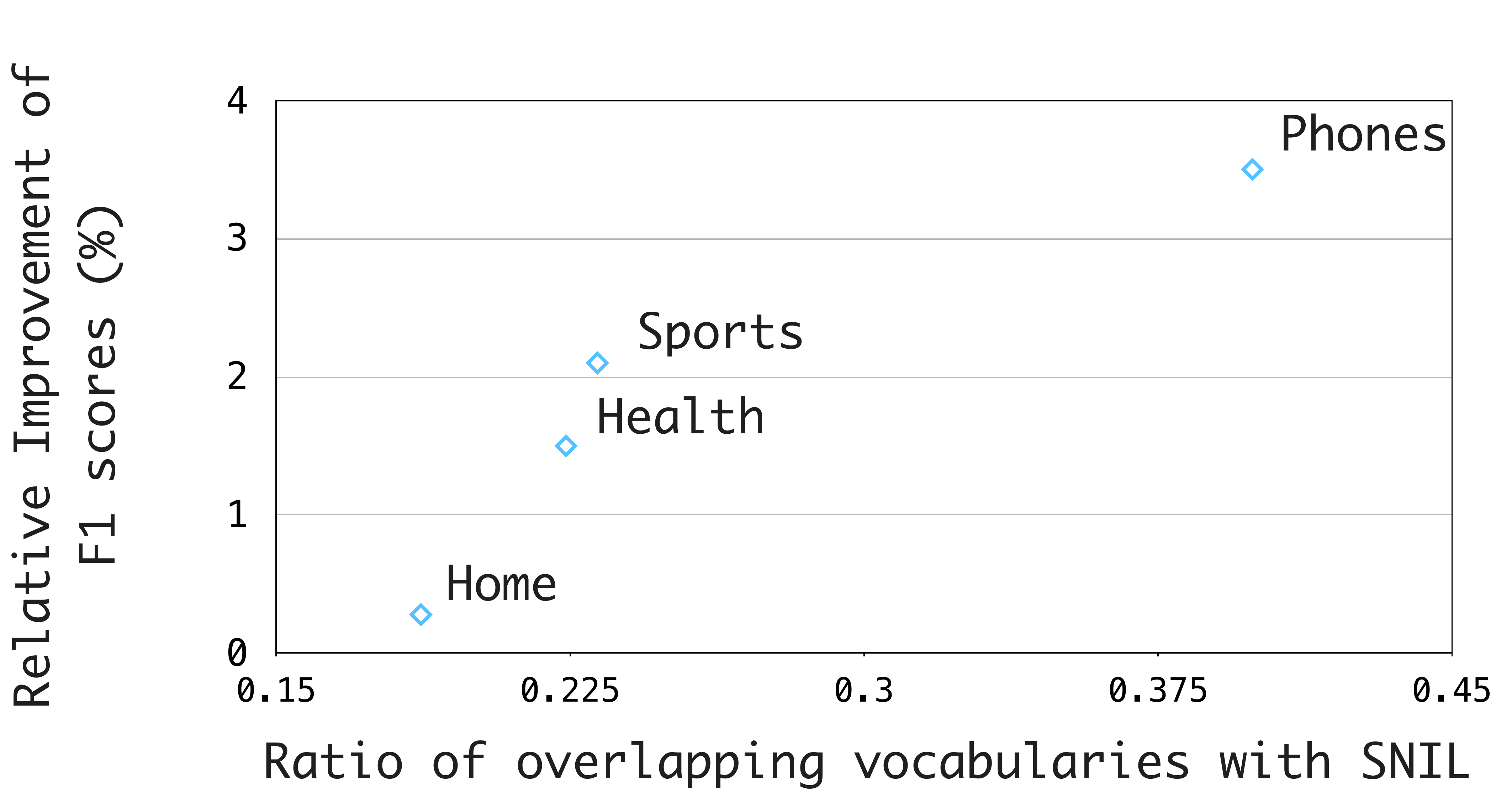}
  \caption{The relationship between the ratio of word overlapping and relative improvement measured by F1 scores} 
  \label{word-overlap}
\vspace{-0.4cm}
\end{figure}

As can be observed from Table \ref{results}, pre-training on the SNLI dataset can help equip the network with some prior knowledge of recognizing review-answer entailment patterns, thus improving the performance. However, the improvements vary from category to category. To gain some insight of what factor causes such difference, we investigate the ratio of the overlapping vocabularies between datasets of several product categories with the SNLI dataset, since the domain difference is often a key factor in transfer learning \cite{conversation-google}. The results are shown in Figure \ref{word-overlap}. We can see that there exists a trend between the overlapping ratio and the relative improvements: larger overlapping ration generally leads to a larger improvement. For example, the performance on the Phone category has been improved for about 3.5\% with the overlapping ratio of 40\%, while the performance of the Home category with the overlapping ratio being 18\% almost does not change. 
This result also suggests that one effective approach for further improving the performance, especially for categories with low performance, can be achieved by providing additional in-domain labeled data of those categories. 

\subsubsection{\textbf{Ablation Study}}

\begin{table}
  \setlength{\abovecaptionskip}{0pt}
  \setlength{\belowcaptionskip}{2pt}
  \fontsize{9}{10}\selectfont
  \caption{Ablation experiments with reported F1 scores}
  \label{tab:ablation}
  \begin{tabular}{l|cccc}
    \toprule
    Models & Sports & Health & Phones & Toys  \\
    \midrule
    RAHP-Base & \textbf{0.655} & \textbf{0.657} & \textbf{0.608} & \textbf{0.633} \\
    - w/o RA coherence &  0.628 & 0.641 & 0.584 & 0.608\\
    - w/o Q-to-R attention &  0.631 & 0.643 & 0.603 & 0.613\\
    - w/o char embedding & 0.640 & 0.653 & 0.600 & 0.627 \\
    \bottomrule
  \end{tabular}
\vspace{-0.4cm}
\end{table}

To investigate the effectiveness of each component in RAHP, we conduct ablation tests by removing different modules of RAHP-Base. Table \ref{tab:ablation} presents the F1 scores of different variant models. The results show that the model without the RA coherence component (RAHP w/o RA coherence) suffers a large decrease, indicating that considering the opinion coherence between the answer and reviews contribute to the final performance. One interesting phenomenon is that different domains suffer different degrees of such performance decrease. For example, the Health domain has a larger degradation compared with other domains. This may due to the fact that questions and answers in this domain are much more subjective and diverse, making the reviews are less discriminative to help identify helpful answers.

Removing the question attention to the reviews (RAHP w/o Q-to-R attention) also leads to a performance decrease, showing the effectiveness of utilizing questions to highlight important concerned information in the reviews.
In addition, discarding character-based embedding from the model (RAHP w/o char embedding) results in performance degradation among all product categories, since the performance may suffer from the common OOV issue caused by misspellings when only using word embeddings.

\subsubsection{\textbf{Case Study}}
\begin{table}
  \small
  \caption{A sample case of multiple answers with original user votes and helpfulness judged by RAHP and ESIM model}
  \label{case}
  \begin{tabular}{p{4.8cm}ccc}
    \toprule
    \multicolumn{4}{c}{\textbf{Product}: \; CHOETECH Wireless Charger} \\ \midrule
    \multicolumn{4}{l}{\textbf{Question}: \; Will it work with a thin or somewhat thin case?} \\ \midrule
    \multicolumn{4}{l}{\textbf{Review Snippets (partial):}} \\
    \multicolumn{4}{l}{"Works well, even with a (thin) case on.";} \\
    \multicolumn{4}{l}{"The CHOE charger works with the phone either bareback or in a case} \\
    \multicolumn{4}{l}{that has a back."} \\
    \midrule 
    \hfil \textbf{Answer} & \textbf{Votes} & \textbf{RAHP} & \textbf{ESIM} \\
    \midrule
    I have a Nexus 7 second gen with a Poetic   & \multirow{2}{*}{[3,3]} & \multirow{2}{*}{Helpful} & \multirow{2}{*}{Helpful}  \\
    case and it charges with no problem!\\
    \midrule
    No, the case you need is too big for any ad-&  \multirow{4}{*}{[0, 4]} & & \multirow{4}{*}{Helpful} \\
     ditiona cases. I do love the product though.&&Not\\
    I have 2 and keep one by my desk and one &&Helpful\\
    on the nightstand. Makes life easier:-) \\
  \bottomrule
\end{tabular}
\vspace{-0.4cm}
\end{table}

To gain some insights of the prediction performance of our proposed model, we present a sample case of multiple answers to a question as shown in Table \ref{case}, including the predicted helpfulness given by RAHP model and a strong baseline ESIM model. We also show their original helpfulness votes as well as some relevant review snippets. 
We can see that both models successfully predict the helpfulness of the first answer, which 3 out of 3 users vote it as a helpful answer, since it mentions a specific case used by that user. However, ESIM model fails to handle the second answer since it does actually talk about whether the concerned product can be used with a case and thus is quite topically relevant. But the information in the reviews further indicates that many customers think this charger works well with a thin case according to their experience. The actual helpfulness votes given by the community also reflect such idea. RAHP utilizes these opinion information and gives a correct prediction of its helpfulness. This real-world case indicates the importance of considering the review information in identifying the answer helpfulness.

\section{Conclusions}
We study a novel task of predicting the answer helpfulness in E-commerce in this paper. To tackle this task, we observe that we need to model both the interactions between QA pairs and the opinion coherence between the answer and common opinions reflected in the reviews. Thus, we propose the Review-guided Answer Helpfulness Prediction (RAHP) model to predict the answer helpfulness. Moreover, a pre-training strategy is employed to help recognize the textual inference patterns between the answer and reviews. Extensive experiments show that our proposed model achieves superior performance on the concerned task. 

\bibliographystyle{ACM-Reference-Format}
\balance
\bibliography{main.bib}

\end{document}